\documentclass[a4paper]{locconf}
\usepackage{graphicx}
\usepackage[utf8]{inputenc}
\usepackage{import}
\usepackage{amsmath}
\usepackage{amsfonts}
\usepackage{amssymb}
\usepackage{hyperref}
\usepackage{url}

\usepackage{afterpage}

\usepackage{xcolor}

\begin{document}

\title{Problems of dataset creation for light source estimation}

\author{E.I. Ershov, A.V. Belokopytov, A.V. Savchik}

\address{Institute for Information Transmission Problems, Russian Academy of Sciences}

\begin{abstract}{Abstract}
The paper describes our experience collecting a new dataset for the light source estimation problem in a single image. 
The analysis of existing color targets is presented along with various technical and scientific aspects essential for data collection. 
The paper also contains an announcement of an upcoming 2-nd International Illumination Estimation Challenge (IEC 2020).
\end{abstract}

\section{Introduction}

Light source estimation is an essential part of the color image formation pipeline in modern mobile devices. 
Currently, the classic and well-established approach to digital camera images visualization includes three necessary stages \cite{brown2016pipeline}: illumination estimation, illumination correction, and conversion to the CIE XYZ color space (CIE standard observer color space) \cite{nikol}. 
The usual assumption is that  CIE XYZ color space leads to a better quality of color reproduction or ``photorealism''. 
However, photorealism as a goal is often abandoned for aesthetics while taking pictures with modern mobile cameras, but the above three stages usually are preserved.

\begin{figure}[h]
    \centering
    \includegraphics[width=0.7\linewidth]{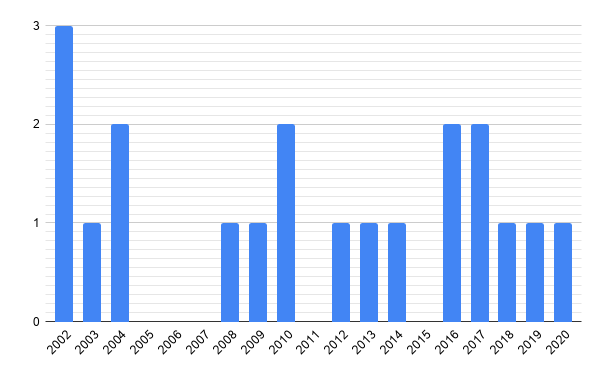}
    \caption{The annual number of open dataset publications on the topic of color constancy.}
    \label{fig:frequency}
\end{figure}

Continued publication of open datasets on this topic is further evidence of the importance of this task. 
Since 2002, on average, there is at least one new dataset every year 
(Fig.~\ref{fig:frequency})\footnote{These statistics were obtained mainly via the network resource \href{https://colorconstancy.com}{colorconstancy.com}}. 
Such activity also indicates that standard methods for assessing the quality of this particular problem solution have not yet been formed in the world scientific community (as happened, for example, in the task of stereo correspondence algorithms evaluation \cite{szeliski2002taxonomy}).


A large IntelTAU dataset was published last year \cite{laakom2019tau}. 
It solves several problems with previously presented datasets. However, it has drawbacks (discussed below) that prevent it from occupying the ``standard'' niche. Another quite large and diverse dataset, also introduced last year, is CubePlus \cite{banic2017unsupervised}. 
This dataset was used for the first international illumination estimation challenge in 2019. 
Nevertheless, as we will see, it is also not perfect.

This paper proposes some meaningful analysis of the existing color targets problems. It lists the technical and scientific difficulties encountered in preparing a new dataset for the 2-nd Illumination Estimation Challenge starting 1-st July 2020, which is organized by the Institute for Information Transmission Problems (IITP RAS) in cooperation with University of Zagreb, Croatia.

We also invite anyone to participate in this competition for scene illumination estimation, see \url{http://chromaticity.iitp.ru}.

\section{Disadvantages of the modern datasets}

Let us consider one of the latest datasets -- IntelTAU \cite{laakom2019tau}.
The authors of this dataset point out its advantages compared to the predecessors.

\begin{itemize}
    \item \textbf{A large number of images.} IntelTAU is considered to be the largest dataset, with just over seven thousand images: three thousand from one of the cameras and two thousand from each of two other cameras.
    \item \textbf{GDPR compatibility.} This dataset meets the requirements of European standards for personal data protection. Images do not contain private information, such as faces and license plates.
    \item \textbf{A variety of cameras.} Images were obtained from three cameras with different sensors, which allows testing the stability of the algorithm to sensor changes.
    \item The authors highlight \textbf{the absence of calibration objects in the frame}. Images with a color target were obtained separately and are almost not represented in the dataset.
\end{itemize}

However, it is difficult to agree with the last point, since the authors provide their own light source estimations for images without color targets. It turns out it is impossible to evaluate the quality of the annotation, and experience shows just how important this verification is: in one of the most popular datasets, markup errors are  found continuously and corrected \cite{hemrit2018rehabilitating} -- the presence of a color target in the frame allows for the human factor influence reduce.

Also, in order to meet the GDPR conditions, the faces were blurred in this dataset. Thus, many images are somewhat artificial, which may affect the use of methods based on machine learning. There are other disadvantages, including those common among many other datasets.

\begin{figure}
    \centering
    \includegraphics[width=0.7\linewidth]{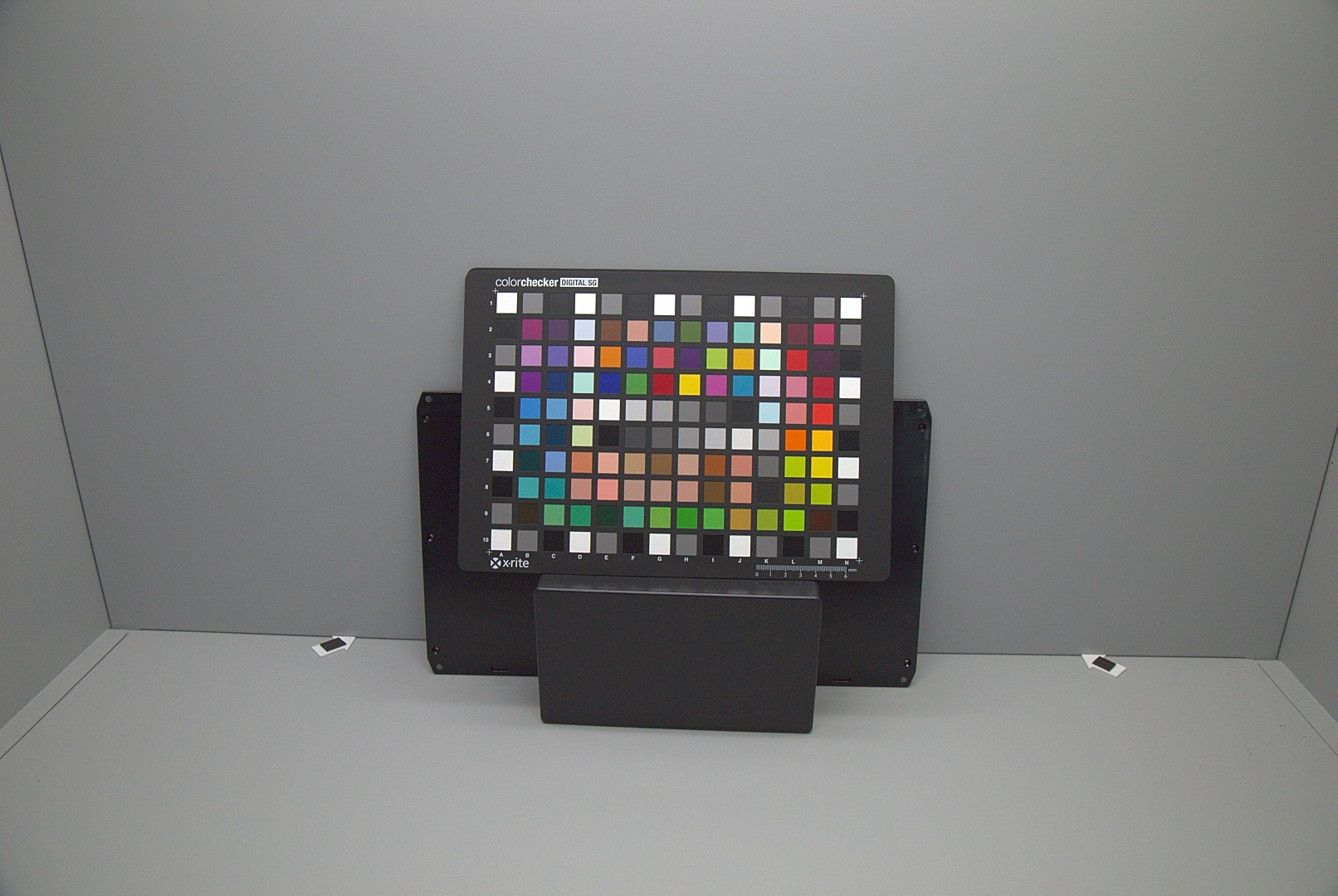}
    \caption{ Example of a flat color target. The image was taken from the IntelTAU dataset.}
    \label{fig:colorchecker}
\end{figure}

Many datasets, including IntelTAU, estimate illumination using a flat color target (Fig.~\ref{fig:colorchecker}).
With a flat color target, it is possible to perform only a single light source estimation, but what kind of estimation will it be, say, on a clear sunny day? 
If the color target is in the shadow (not illuminated by the sun), the only light source will be the sky,
which is usually a significantly less bright light source than the sun. 
If the color target is not in the shadow, then the sun and sky are light sources acting simultaneously.
Hence, the mixture of two sources is estimated. 
As a result, the algorithms and models trained on such datasets solve quite an incorrect problem
for images taken on a sunny day: to predict a precise linear mixture of colors of two light sources with unknown proportions.
This drawback can only be eliminated with the help of the three-dimensional color targets,
as was done, for example, in the \cite{ciurea2003large} dataset.
One good modern example of such a dataset is CubePlus \cite{banic2017unsupervised}).

In addition to the above problems, there are three more drawbacks that are characteristic of all existing datasets.

\begin{enumerate}
    \item \textit{No additional annotation.}
    Related information on the type of scene, time of day, the presence of objects of known color, and other semantic features are valuable for color constancy algorithms. Such information could allow not only to calculate the aggregated value -- the quality of the solution for all images but also to analyze the error clusters. For example, a specific light source estimation algorithm can work well for images shot in natural daylight conditions but not in artificial lighting.
    \item \textit{Insufficient number of images.} 
    Even a record-setting IntelTAU dataset contains no more than three thousand photos per sensor, which often contains repeated images of the same scene; moreover, this applies to older datasets as well. 
    For example, a classic dataset for light source estimation, ColorChecker \cite{hemrit2018rehabilitating}, has less than a thousand images. 
    Such a limited quantity may not be enough to develop color constancy algorithms designed for various types of scenes. 
    \item \textit{Incompatibility of datasets.} 
    Each subsequent dataset is often collected using a camera different from the ones used to collect previous datasets, and color targets may also differ.  
    Thus, it is problematic to combine the accumulated datasets into one. 
    In this situation, it is difficult to analyze the source of the algorithm error -- 
    was it a complexity of the scene, or was it because the camera's sensitivity characteristics were different from those used at the algorithm development stage?
    Rare examples of consistent dataset additions are CubePlus, which expands the Cube dataset, and IntelTAU, which expands IntelTUT.
\end{enumerate}

A possible solution to the above problems is to substantially expand and annotate one of the large modern datasets containing a three-dimensional color target.
The authors adhere to this approach using the CubePlus dataset as a basis.

\section{Color targets}

Most modern and well-known datasets use flat color targets of different manufacturers, such datasets include: NUS \cite{cheng2014illuminant}, REC \cite{hemrit2018rehabilitating}, INTEL-TUT \cite{aytekin2019tut}, INTEL-TAU \cite{laakom2019tau}, MLS \cite{sma}.

As mentioned above, it is impossible to correctly estimate light source using a flat color target if two significant sources of illumination are present in the scene (note: this applies not only to photographs taken on a clear day but also to a multitude of images with various artificial light sources taken in the night street as well as indoors). 
On the other hand, the advantage of color targets is that they are suitable not only for the evaluation of illumination estimation accuracy, but also for the quality evaluation of problem solution of color correction, and for a transition to the XYZ color space. 
It should also be noted that specifically flat color targets are usually manufactured with the highest standards of precision.

Among three-dimensional color targets that were used in datasets, only two can be listed: a gray ball (as in the GrayBall \cite{ciurea2003large} dataset) and SpyderCube (CubePlus \cite{banic2017unsupervised}).
It is quite possible to use several flat color targets or flat color targets, which are made foldable for compactness, to reach 3D quality.
However, such public datasets are currently unknown to the authors.

Generally, a gray ball seems to be a better color target than SpyderCube, since it allows a more detailed illumination estimation. 
First, by analyzing the histogram of the pixels' colors in the image of a gray ball, it is possible to estimate the number of sources (illumination rank) directly, if there are no more than three of them. 
Specifically, if non-zero values on the histogram fit into one straight line, then the scene has one source, if it fits on the plane -- two, if the entire histogram is densely filled -- then there are three sources. 
Secondly, in addition to the histogram analysis, the shadows on this ball can be considered, which allows increasing the maximum illumination rank, which can be estimated \cite{bouganis2004multiple}.

Despite the above, choosing a gray ball as a color target currently is not the best solution. The existing dataset collected in 2003 is already outdated, and color targets in the form of a gray ball with stable color characteristics are not available on the market.

\begin{figure}[h]
    \centering
    \includegraphics[width=0.7\textwidth]{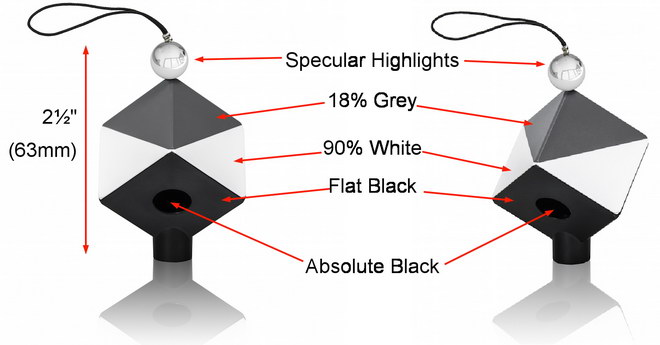}
    \caption{SpyderCube color target by DataColor\protect\footnotemark.}
    \label{fig:spydercube}
\end{figure}

\footnotetext{Image from \href{https://www.displaydaily.com/index.php?option=com_content&view=article&id=50684:cinematography-hdr-and-waveform-monitors}{displaydaily.com}}

Hence, it is advisable to use Datacolor's SpyderCube (Fig.~\ref{fig:spydercube}) to create the most representative and largest dataset among existing ones  -- about 2000 images have already been collected with it, and this target makes it possible to determine the plane on the color histogram that contains the sources' colors. 
Besides, SpyderCube can be used in a wide range of exposures: 
there are white areas on the edges for dark scenes estimation, 
and gray areas for bright scenes. 
It also contains an imitation of a completely black body (a hole in the base), that can be used to estimate the camera's black level. 
Finally, the mirror ball at the top contains information about the world above the scene in a compressed form, including information about other light sources.

\section{Notes on the dataset collection in uncontrolled conditions}

\begin{figure}[h]
    \centering
    \includegraphics[width=0.7\linewidth]{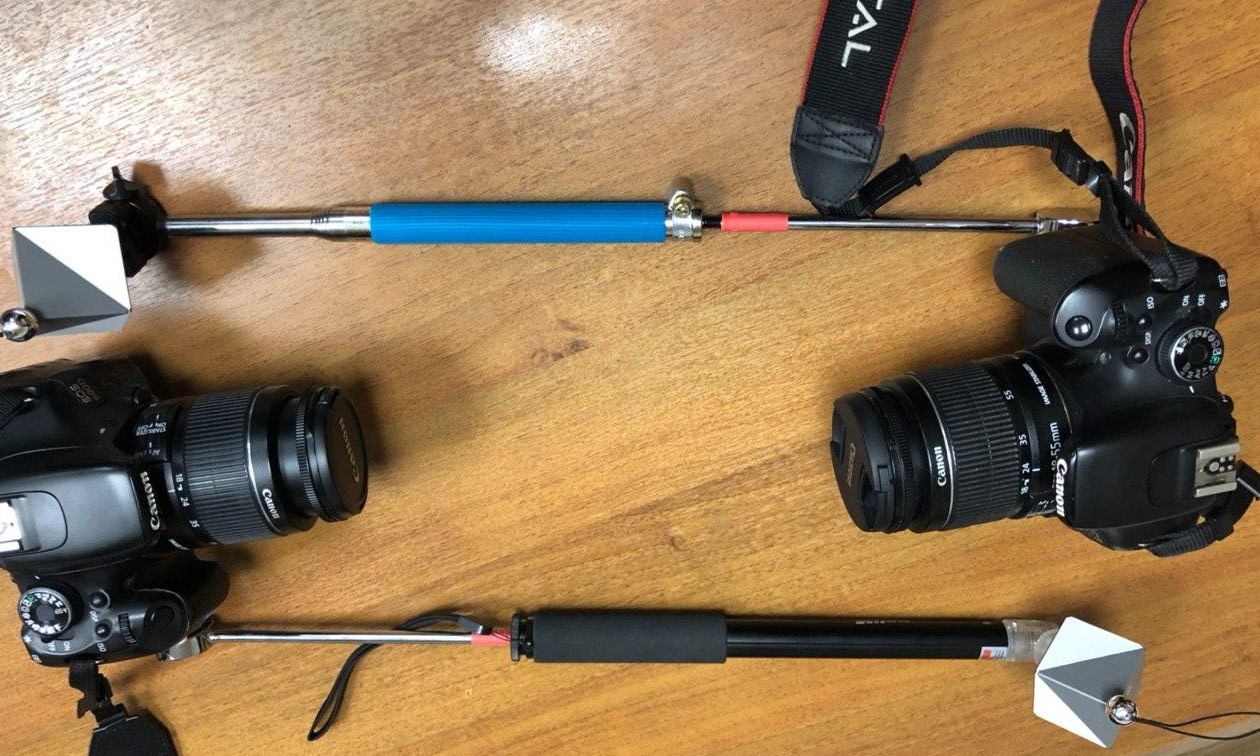}
    \caption{Two handheld setups for collecting data in uncontrolled conditions.}
    
    \label{fig:mobile}
\end{figure}

We created two identical handheld setups to collect the dataset in natural (uncontrolled) conditions.
Cameras with color targets mounted on them are shown in Fig.~\ref{fig:mobile}.

\begin{figure}[h]
\center
\begin{minipage}[h]{0.23\linewidth}
\center{\includegraphics[width=1\linewidth]{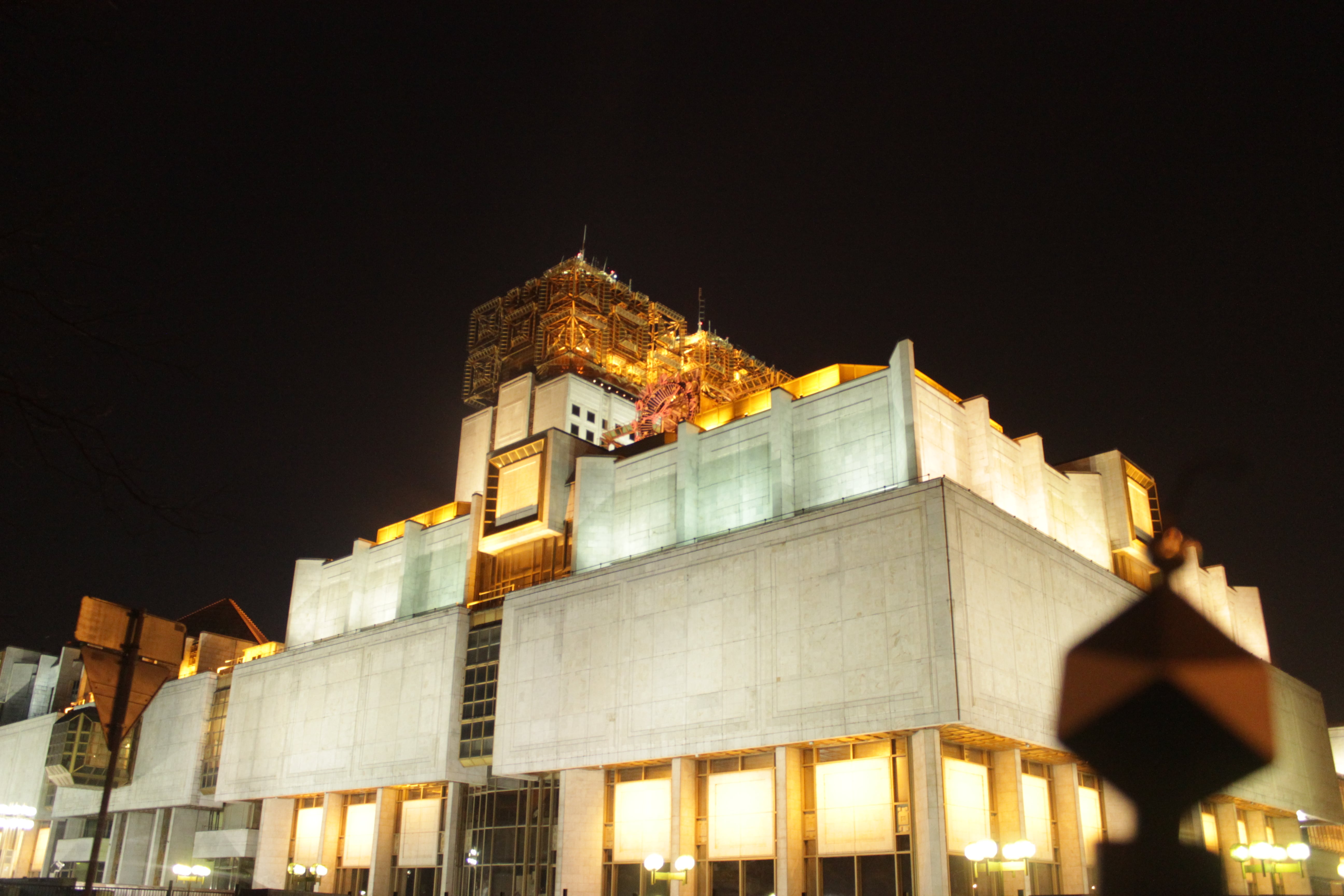} \\ (a)}
\end{minipage}
\begin{minipage}[h]{0.23\linewidth}
\center{\includegraphics[width=1\linewidth]{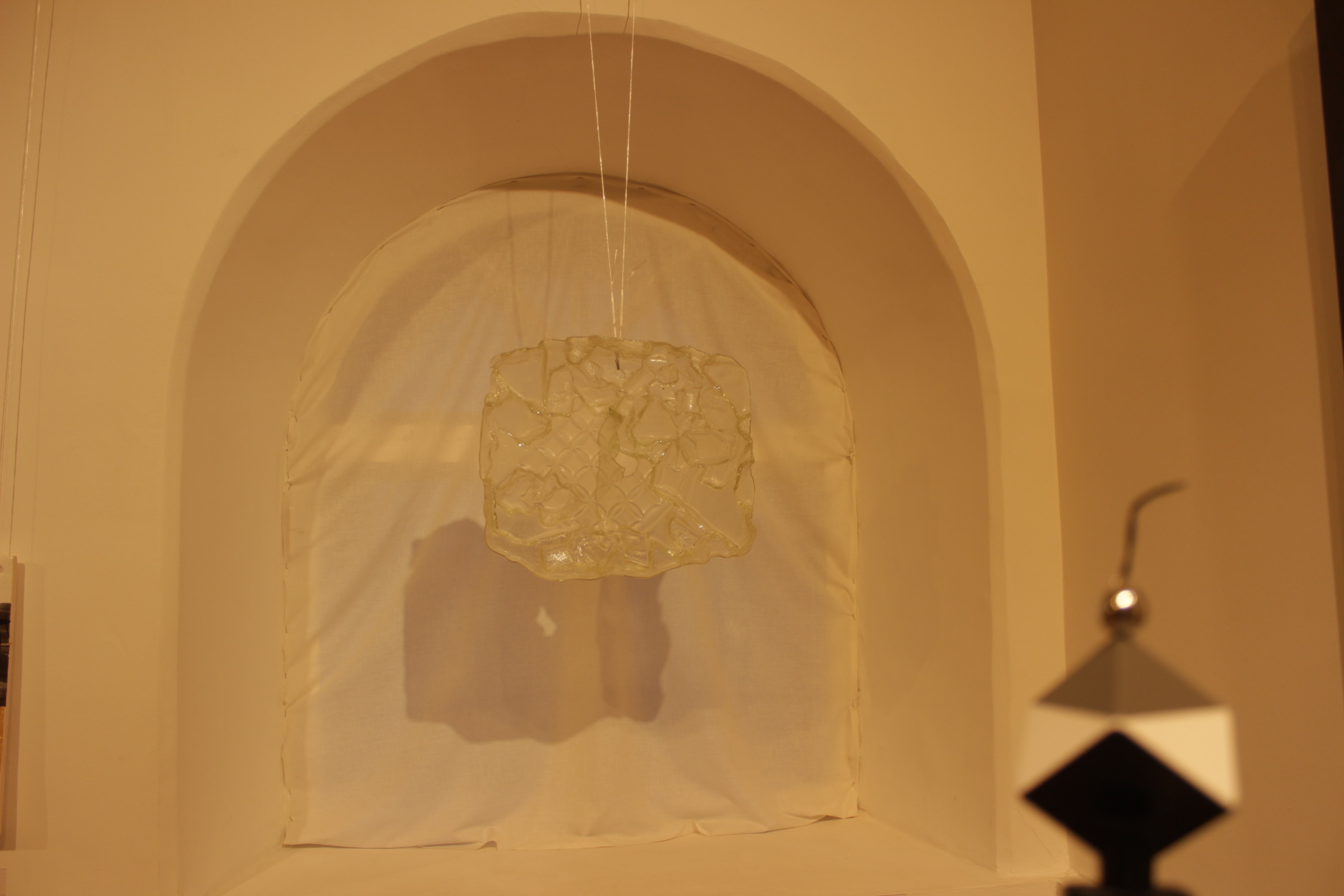} \\ (b)}
\end{minipage}
\begin{minipage}[h]{0.23\linewidth}
\center{\includegraphics[width=1\linewidth]{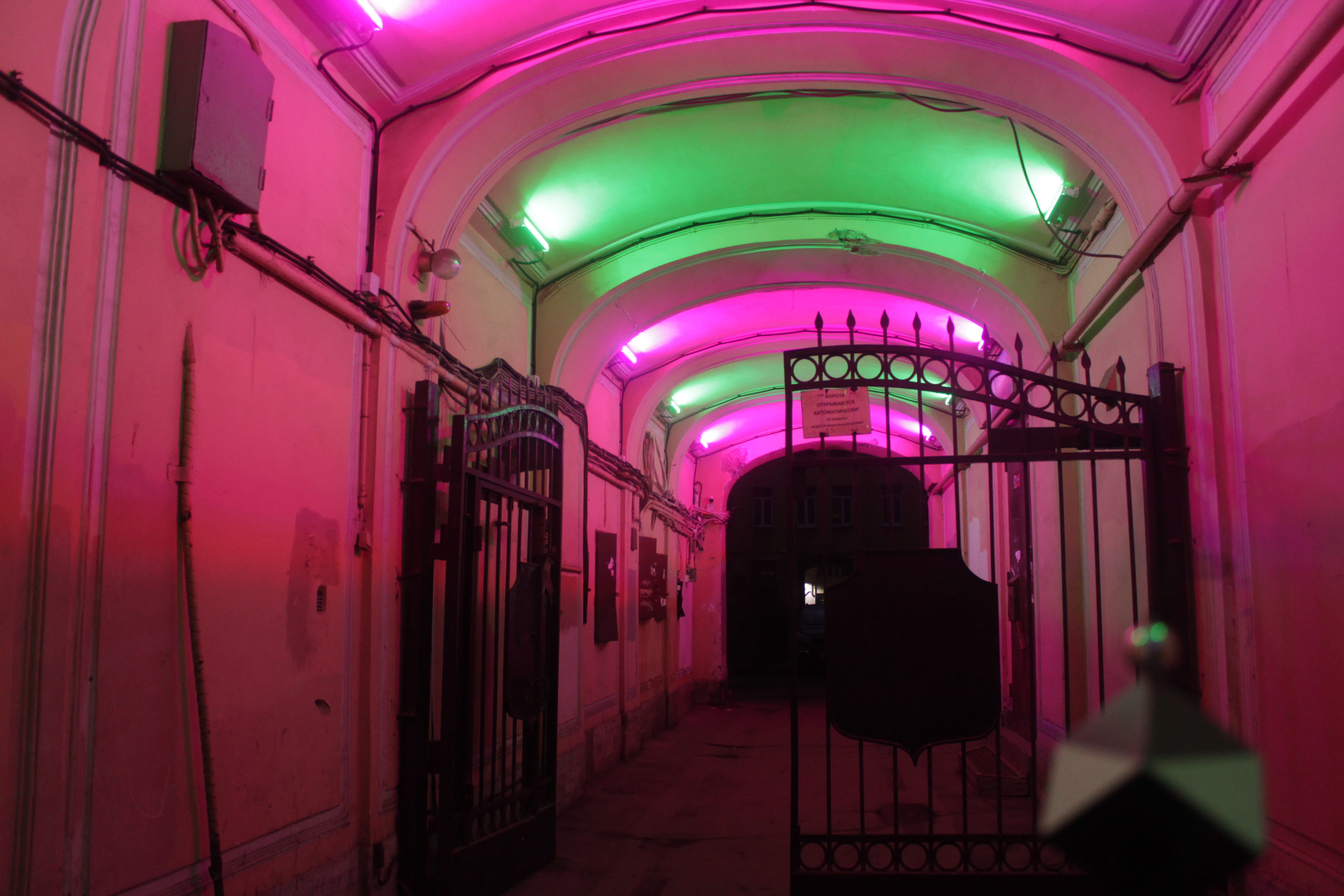} \\ (c)}
\end{minipage}
\begin{minipage}[h]{0.23\linewidth}
\center{\includegraphics[width=1\linewidth]{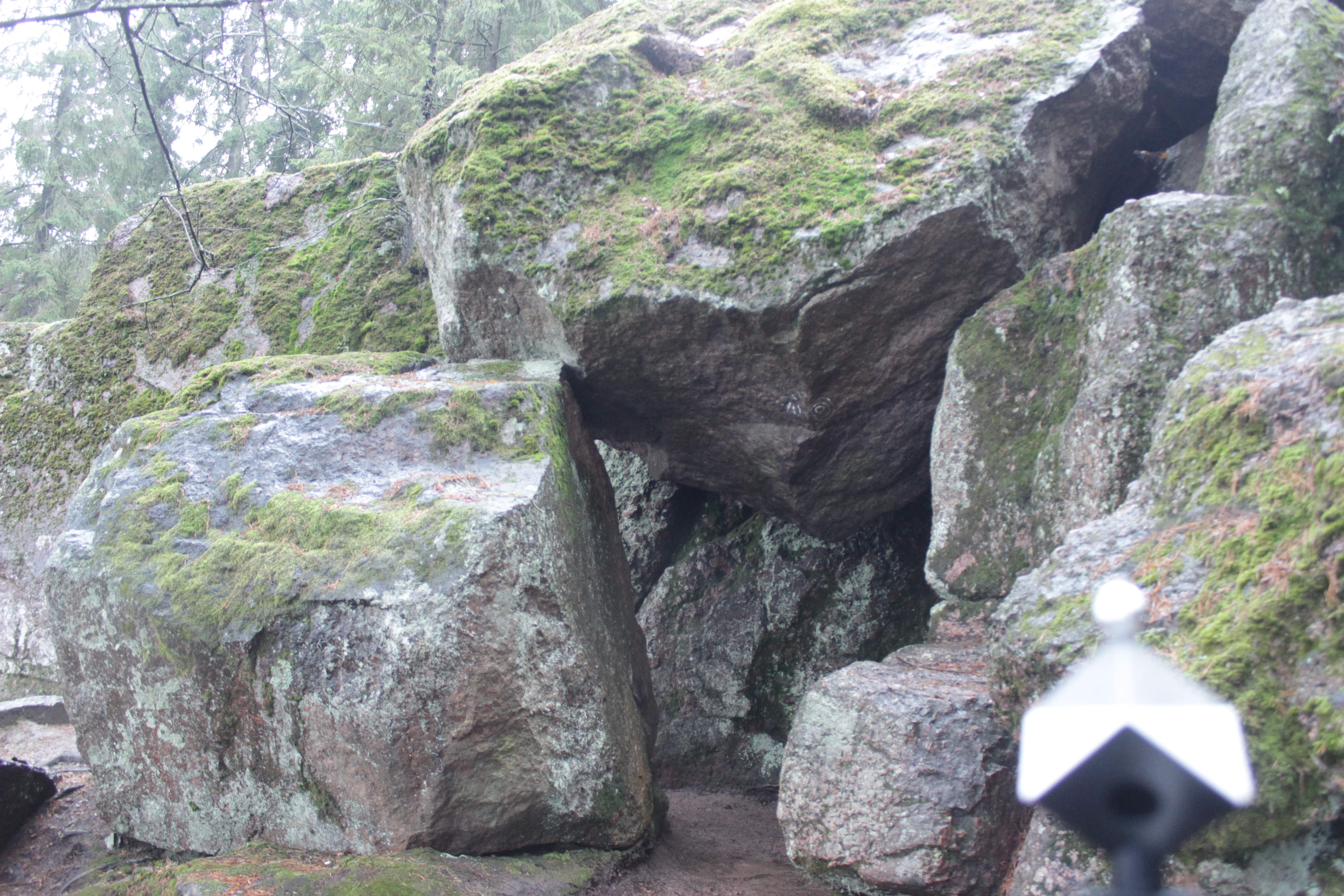} \\ (d)}
\end{minipage}

\caption{
Examples of photographs that should be excluded from the dataset: 
a) the color target is illuminated by a lantern, the color of which is different from the lighting of the building;
b) the lighting of the cube's borders is the result of multiple reflections in a room with almost white walls;
c) the color target is illuminated by sources that have almost no effect on the lighting of the observed scene;
d) the majority of the color target pixels are clipped white due to overexposure.
\label{fig:examples}
}
\end{figure}

One of the key factors in the dataset collection is the consistency between the illumination of the color target and the observed scene.
The same source (or a pair of sources) must illuminate both the entire scene and the color target.
Otherwise, the source color estimation will be irrelevant to the most pixels of the scene.
Such situations occur when the local illumination near the photographer does not match the illumination of the scene being shot 
(examples are shown in Fig.~\ref{fig:examples}a-c).

Another important factor is the presence of glare on the color target. 
For those images where for dark gray faces of the color target there are clipped values in one of the channels, it is impossible to extract a correct illumination estimation (an example is presented in Fig.~5d). 
To avoid such images, one should consider either manual camera settings or specifying relative exposure compensation to camera automatic estimation. 
In our case, dimming by one step is usually enough. 
Separately, we should note the importance of correct photography of scenes with two light sources. 
It is highly recommended, especially if the image areas of the scene illuminated by different light sources are comparable, place the color target so that the colors of its different faces noticeably differ. In this case, it is possible to estimate the plane in the color space containing the colors of the left and right sides, thereby the evaluation of two significant light sources is controlled at once.

As shown, additional semantic markup of the scene content is valuable. So we propose the following feature hierarchy:
\begin{enumerate}
    \item Time of day (day/night/unknown)
    \item Location (indoor/outdoor)
    \item The presence of light source in the frame (sun/sky/lamps/street lights)
    \item The presence of objects that can be identified (human faces/license plates, etc.)
    \item The presence of objects of known color in the scene
    \item The presence of shadows (yes/no)
    \item The richness of the scene (rich/poor/unclear) 
    \item Main lighting (sun/sky/lamp/lantern/unclear)
\end{enumerate}

The hierarchy is incomplete, but currently, it can profile various illumination estimation algorithms and identify challenging scenarios.

\section{Conclusion}

In this article, we discuss the main pros and cons of the largest datasets available for the light source estimation problem. We present a brief analysis of three-dimensional color targets and highlight some of the technical and scientific details that we discovered when collecting our new dataset. We provide typical examples of images with the incorrectly illuminated color target and, finally, suggest relevant image categorization.

The collected dataset will be used in the upcoming 2-nd Illumination Estimation Challenge (IEC 2020) starting 1-st July 2020 and organized by the Institute for Information Transmission Problems (IITP RAS) in cooperation with University of Zagreb, see \url{http://chromaticity.iitp.ru}.

We are pleased to invite all interested teams to participate in the competition.

\smallskip

\end{document}